\documentclass[10pt,twocolumn,letterpaper]{article}

\usepackage{wacv}
\usepackage{times}
\usepackage{epsfig}
\usepackage{graphicx}
\usepackage{amsmath}
\usepackage{amssymb}

\usepackage{booktabs}

\def\wacvPaperID{109} % Enter the WACV Paper ID here

\wacvfinalcopy % *** Uncomment this line for the final submission

\ifwacvfinal
\fi

\ifwacvfinal
\usepackage[breaklinks=true,bookmarks=false]{hyperref}
\else
\usepackage[pagebackref=true,breaklinks=true,colorlinks,bookmarks=false]{hyperref}
\fi

\def\ie{\emph{i.e.,}} 
 
\def\etc{\emph{etc.}}

\def\Vec#1{{\boldsymbol{#1}}}
\def\Mat#1{{\boldsymbol{#1}}}

\newcommand{\fig}{Fig.}
\newcommand{\tab}{Table}

\begin{document}

%%%%%%%%% TITLE
\title{Saliency Prediction with External Knowledge}

\author{Yifeng Zhang, Ming Jiang, Qi Zhao\\
University of Minnesota, Twin Cities\\
%Kenneth H. Keller Hall, 200 Union St SE, Minneapolis, MN 55455\\
%{\tt\small zhan6987@umn.edu},
% For a paper whose authors are all at the same institution,
% omit the following lines up until the closing ``}''.
% Additional authors and addresses can be added with ``\and'',
% just like the second author.
% To save space, use either the email address or home page, not both
%{\tt\small mjiang@umn.edu},
%{\tt\small qzhao@cse.umn.edu}
}

% set the model name
\newcommand{\model}{GraSSNet}
\newcommand{\modelfullname}{Graph Semantic Saliency Network}

\maketitle
%\thispagestyle{empty}

% % set starting page
% \begin{verbatim}
% \setcounter{page}{1}
% \end{verbatim}

%%%%%%%%% ABSTRACT
\begin{abstract}
The last decades have seen great progress in saliency prediction, with the success of deep neural networks that are able to encode high-level semantics. 
Yet, while humans have the innate capability in leveraging their knowledge to decide where to look (\eg~people pay more attention to familiar faces such as celebrities),  
saliency prediction models have only been trained with large eye-tracking datasets.
This work proposes to bridge this gap by explicitly incorporating external knowledge for saliency models as humans do. We develop networks that learn to highlight regions by incorporating prior knowledge of semantic relationships, be it general or domain-specific, depending on the task of interest.
At the core of the method is a new \modelfullname~(\model) that constructs a graph that encodes semantic relationships learned from external knowledge. A Spatial Graph Attention Network is then developed to update saliency features based on the learned graph. Experiments show that the proposed model learns to predict saliency from the external knowledge and outperforms the state-of-the-art on four saliency benchmarks.
\end{abstract}

%%%%%%%%% BODY TEXT % 8 pages in total to be trimmed
\section{Introduction}
Visual attention is the ability to select the most relevant part of the visual input. It helps humans to rapidly process the overwhelming amount of visual information acquired from the environments. Saliency prediction is a computational task that models the visual attention driven by the visual input~\cite{itti2001computational}, which has wide applicability in different domains, such as image quality assessment~\cite{zhang2015application}, robot navigation~\cite{chang2010mobile} and video surveillance~\cite{mancas2011abnormal}, and screening neurological disorders~\cite{jiang2017learning,wang2016revealing,wang2015saliency}.

\begin{figure}
\centering
\includegraphics[width=\linewidth]{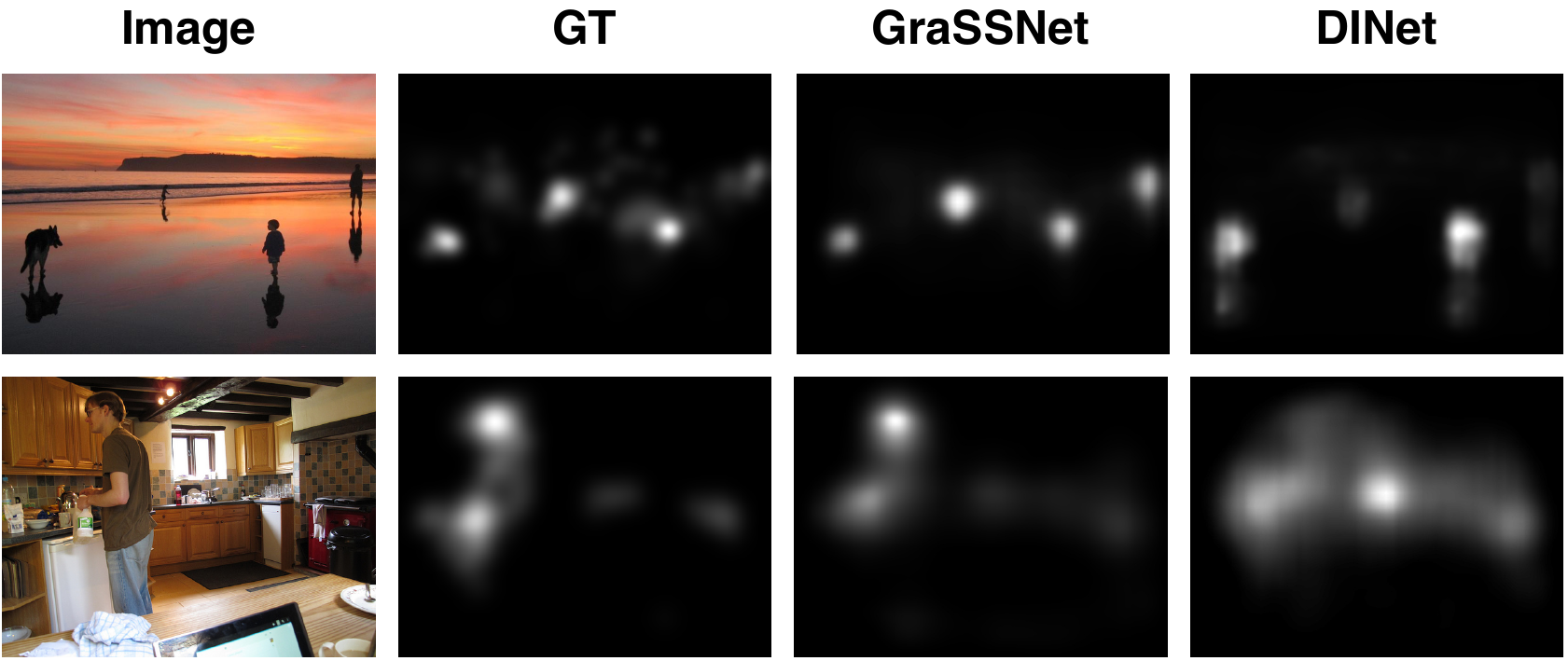}
\caption{\textit{Examples of how semantic proximity affects saliency maps.} External semantic relationships are effective in deciding the relative saliency of objects
}
\label{fig:intro}
\vspace{-0.5em}
\end{figure}

Where humans look is involuntarily influenced by their prior knowledge. Such knowledge can be general commonsense knowledge or specific ones that require prior experience or training~\cite{colonius2011computing,humphrey2009domain,le2009we}. It is commonly noticed that salient objects tend to influence the saliency of similar objects. For example, as illustrated in \fig~\ref{fig:intro}, when multiple people and objects exist, their saliency values relevant to the closeness of their relationships. When one of them is salient, their related objects also tend to be salient.
% For example, as illustrated in \fig~\ref{fig:intro}a, we know that when people, food, and kitchenware coexist in a scene, commonly kitchen, dining room, or restaurant, these objects are relevant and interesting. The commonsense relationship between food and people helps boost their importance. In contrast, \fig~\ref{fig:intro}b shows that food without people attracts less attention.

Differently, despite the success of deep neural networks for saliency prediction~\cite{bruce2016deeper,huang2015salicon,kruthiventi2016saliency}, they rely on training data and learn `knowledge' only in a data-driven and implicit manner. With the advancement of DNNs and the collection of more data, these networks learn better semantics that encode objects and maybe high-level context or relationship; it is, however, unclear to what degree what knowledge can be learned, and it heavily depends on the data quantity and content. Therefore, we in this work propose to leverage ground truth knowledge from external sources. Such knowledge could well complement the features learned from the neural networks to more intelligently decide where to look. Note that attention data are not trivial to scale~\cite{nguyen2018attentive}, which makes this work more useful in practice, \eg~domain-specific knowledge could be directly used to guide saliency prediction in a clinical application without big attention ground truth.  
 
To demonstrate the overarching goal, we use two external knowledge sources (MSCOCO image captioning~\cite{lin2014microsoft} and WordNet~\cite{miller1995wordnet}) that describe semantic relationships between objects. Semantic relationship is important to saliency prediction as objects are correlated and together they also reflect context; it is also one of the most well-structured and documented sources of external knowledge. We introduce this knowledge into a computational saliency model by designing a \modelfullname~(\model), which explicitly analyzes the semantic relationships of objects in a scene and uses such knowledge for saliency prediction. In particular, we propose a Semantic Proximity Network (SPN) that computes the semantic proximity of detected region proposals in semantic spaces of interest. While external knowledge is explicitly used to supervise the learning of the network, the relationships to be distilled is dependent on the input image by setting the distillation loss as a part of the objective.
We further propose a Spatial Graph Attention Network (sGAT) to propagate the semantic features of region proposals based on their semantic proximity with maintained latent spatial structures, where the updated features will be used together with the multi-scale spatial feature maps to compute saliency maps.

In sum, we propose to explicitly leverage external knowledge for saliency prediction, as a complementary source of information to neural network based models. Extensive experiments on four datasets with comparisons with six models and analyses demonstrate the advantage of incorporating the knowledge. The main technical contributions are summarized as follows:
\begin{itemize}
\item We propose a new graph-based saliency prediction model by leveraging object-level semantics and their relationships.
\item We propose a novel Semantic Proximity Network to explicitly distill semantic proximity from multiple external knowledge bases for saliency prediction.
\item We propose Spatial Graph Attention Network to dynamically propagate semantic features across objects for the prediction of the saliency across multiple objects.
%\item The proposed approach achieves state-of-the-art saliency prediction performance on public visual saliency datasets.
\end{itemize}

% -------------------------------------------------- Relaed Works -------------------------------------------
\section{Related Works}
In this section, we first review state-of-the-art visual saliency models. Next, we briefly introduce how external knowledge is utilized in other high-level computer vision tasks (\eg~relationship detection) and how we adapt it to predict saliency. Lastly, we review and compare graph convolution methods with ours.

\subsection{Deep Saliency Prediction Models}
The recent success of deep learning models has brought considerable improvements in saliency prediction tasks. One of the first work is Ensemble of Deep Networks (eDN)~\cite{vig2014large}, which combines multiple features from a few convolution layers. Later DeepGaze I~\cite{kummerer2014deep} leverages a deeper structure for better feature extraction. After that, many models~\cite{jia2020eml,kroner2019contextual,kruthiventi2017deepfix} follow the framework that consists of a deep model and fully convolutional networks (FCN) to leverage the powerful capabilities in contextual feature extraction. These models are often pre-trained on large datasets (\eg~SALICON~\cite{jiang2015salicon}) and then fine-tuned on small-scale fixation datasets. However, with the increasing model depth, many downsampling operations are performed, contributing to a lower spatial resolution and limited performance~\cite{liu2018deep}. A recent state-of-the-art model named Dilated Inception Networks (DINet)~\cite{yang2019dilated} leverages dilated convolutions to tackle the issue. 
Another major strength of the deep model is its capabilities of high-level feature extraction. Many deep neural network based method \cite{bruce2016deeper,huang2015salicon,jiang2015salicon,kruthiventi2016saliency} have boosted saliency prediction performance by implicitly encoding semantics with different approaches (\eg~subnets in different scales~\cite{huang2015salicon}, inception blocks~\cite{szegedy2015going}, \etc). 

However, none of the previous methods explored the saliency patterns among different objects in a scene. Our model differentiates itself from existing methods by leveraging the relationships of various semantics for saliency prediction, using a Semantic Proximity Network and a Spatial Graph Attention Network. 

\subsection{External Knowledge Distillation}
External knowledge has gained great interest in natural language processing~\cite{bao2014knowledge,hinton2015distilling} and computer vision~\cite{aditya2018explicit,deng2014large,li2017incorporating}. As the information extracted from training sets are always insufficient to fully recover the real knowledge domain, previous works explicitly incorporate external knowledge to compensate it. Generally, there are two commonly used frameworks for knowledge distillation. 

One framework is \textit{teacher-student distillation}. For example, Yu~\etal~\cite{yu2017visual} leverages this structure to absorb linguistic knowledge in visual relationship detection tasks. Apart from the teacher-student structure, more existing works in object/relationship detection and scene graph generation adopt the \textit{graph framework}. For instance, KG-GAN~\cite{gu2019scene} improves the performance of scene graph generation by effectively propagating contextual information across the external knowledge graph. Similarly, ~\cite{jiang2018hybrid} adopts external knowledge graphs to solve the long-tail problems in object detection tasks. 

Our model also employs a graph framework to distill external knowledge. However, unlike object/relationship detection tasks, where semantics are explicitly defined and structured (\eg~objects, relationships), semantics in saliency feature maps are always entangled, making them non-trivial to connect with external knowledge. To tackle this problem, we not only segment semantics by extracting region proposals, but also convert the external knowledge to image specific region-to-region semantic proximity graphs.

% and , and use external knowledge graphs to supervise the modeling of this graph.

\subsection{Graph Convolution Networks}
Leveraging graphs in saliency prediction has been explored at the pixel level. GBVS~\cite{harel2007graph} treats every pixel as node and diffuses its saliency information along the edges by Markov chains. Recently, graph convolution networks have been applied in various tasks that require information propagation. These methods can largely be categorized into spectral~\cite{bruna2013spectral,henaff2015deep,kipf2016semi} and non-spectral~\cite{atwood2016diffusion,duvenaud2015convolutional,hamilton2017inductive} approaches. One recent approach named Graph Attention Network (GAT)\cite{velivckovic2017graph} achieves state-of-the-art by leveraging self-attention mechanism. 

Inspired by the GAT, we develop a Spatial Graph Attention Network (sGAT) to process spatial feature maps as node attributes. While SGAT assumes no spatial structure within node attributes, our proposed sGAT encodes spatial characteristics during feature propagation, because of their importance in predicting the spatial distribution of attention.

% Compared with GAT, when node features are shared with neighbors, the spatial graph attention network in \model replaces the fully-connected layers with convolutions.

% ------------------------------------------------ Methodology -----------------------------------------
\section{Method}

\begin{figure*}[t]
\centering
\includegraphics[width=1\linewidth]{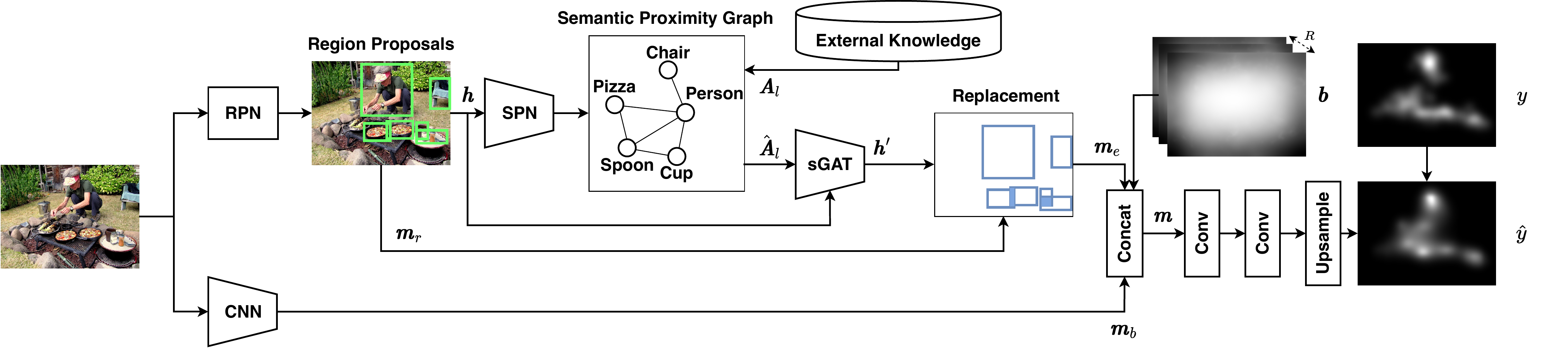}
\caption{\textit{Architecture of the proposed graph semantic saliency network.} The architecture mainly consists of a Region Proposal Network (RPN), a Semantic Proximity Network (SPN), a Spatial Graph Attention Network (sGAT), and a baseline saliency network. The concatenated features from external knowledge (top), baseline saliency network (bottom) and prior maps (optional) are fed into several convolutional and upsampling layers to compute saliency maps}
\label{fig:modelStructure}
\vspace{-0.5em}
\end{figure*}

This section presents the \modelfullname~(\model), as shown in \fig~\ref{fig:modelStructure}. The task
is formulated as follows: given a 2D image $I$ as the input, it aims to construct a semantic proximity graph and use it to predict a saliency map as a 2D probability distribution of eye fixations. We will first describe our model architecture, followed by the details of the two novel components: Semantic Proximity Network (SPN) and Spatial Graph Attention Network (sGAT). Finally, we present the objective function to optimize our model.
% ------------------------------------------------- Model Structures -------------------------------------
% figure ------------------------------------------------------------------------------------

\subsection{Model Architecture}
% is composed of five components, where 
% (1) a baseline saliency network based on spatial convolution, (2) a region proposal network (RPN) that extracts regional features from detected objects~\cite{ren2015faster}, (3) a novel Semantic Proximity Network (SPN) that constructs a semantic proximity graph of the region proposals, and (4) a Spatial Graph Attention Network (sGAT) to compute the updated regional features, (5) the updated regional features are mapped back to their original positions in the spatial feature maps, and concatenated with the feature maps from the baseline saliency network for saliency prediction. Generally, RPN, SPN and sGAT cooperates to incorporate external semantic knowledge into saliency features maps, while the base line saliency network can complement for the necessary features that are omitted during the distillation of external knowledge. With the fusion of internal and external saliency features, our model can better predict saliency maps.

\emph{Object Feature Retrieval.}
Our method is based on detected region proposals. As shown in \fig~\ref{fig:modelStructure}, the model uses a pre-trained Faster R-CNN~\cite{ren2015faster} to detect all objects from the input image $I$, generating a set of bounding boxes $B = \{b_1, \cdots, b_p\}$ where  $p$ denotes the total number of detected instances. Their corresponding regional features $\Vec{h} = \{h_1, h_2, \cdots, h_p\}$, $h_i \in \mathbb{R}^{d_1\times d_2 \times d_3}$  are extracted from the outputs of the ROI pooling layer, where $d_1$, $d_2$ and $d_3$ denote the dimensions of features. %Note that {\color{blue}we do not predict the object categories to diminish the semantic discrepancy impacts of training datasets between RPN and the rest components}.

% It is not possible to explicitly distill object level semantics until objects are detached from the image.

% we also adopt a background network of the state-of-the-art model DINet~\cite{yang2019dilated} to produce background saliency features $\Vec{m}_b$, as some essential information may be omitted during the process of introducing semantic proximity information. 

% The model also includes two other components, the backbone and the background network. The backbone is used to extract rich saliency features from raw image. Apart from that, 

% under the supervision of different external knowledge sources, which serves as the process of incorporating external semantics. To leverage the semantics, regional (object) features are convoluted on the aforementioned graphs. Besides, features from backbone are also fed into background generator for the purpose of creating a background feature map. The background features are concatenated with semantic-rich region features for saliency map synthesis.

\emph{Semantic Proximity Graph Construction.}
To incorporate external knowledge from multiple sources, we process these regional features with a set of $N$ Semantic Proximity Networks (SPNs) that predict the semantic proximity graphs under the supervision from $N$ different external knowledge sources. This design makes it flexible to extend the model with additional knowledge bases. Given regional features $\Vec{h}$, a semantic proximity graph is computed as
\begin{align}
    \hat{\Vec{A}}_l = f_{\text{SPN}}^{l}(\Vec{h}),
\end{align}
where $f_{\text{SPN}}^{l}$ denotes the SPN supervised by an external knowledge graph $\Mat{A}_l$ where $l = 1,\cdots, N$.

\emph{Semantic Proximity Knowledge Distillation.}
Upon obtaining $N$ predicted graphs $\hat{\Mat{A}}=\{\hat{\Vec{A}}_1, \hat{\Vec{A}}_1,\cdots, 
\hat{\Vec{A}}_N\}$, the regional features are processed with $N$ different Graph Attention Networks (sGATs), sharing the saliency features to their immediate neighbors in the  corresponding semantic proximity graphs to generate updated regional features $\Vec{h}'_l=\{h'_{l1}, h'_{l2}, \cdots, h'_{lp}\}$:
\begin{align}
    \Vec{h}'_l = f_{\text{sGAT}}^{\hat{\Mat{A}_l}}(\Vec{h}).
\end{align}
By supervising the SPNs with different externally built ground-truth proximity graphs, diverse proximity knowledge can be learned from external knowledge, so that features can be propagated along the predicted graphs.

% Since we have $N$ predicted graphs $\{\hat{\Vec{A}}_l$, we can obtain $N$ sets of updated regional features
We concatenate all the updated regional features  $\{\Vec{h}'_l\},\, l\in[1, 2, \cdots, N]$ and use a convolution layer to compute the final updated features $\Mat{h}'$, which are projected back to replace the raw map features $\Vec{m}_r$ in the detected bounding boxes. Features in overlapping regions are merged with the max operation.

%. For the overlapped regions, we apply a max operation to keep the most activated features at each pixel. In this way, we obtain the proximity feature maps $\Vec{m}_e$. 
\emph{Prior Maps Generation.} As fixations tend to be biased towards the center of the image~\cite{tatler2007central}, we model this center bias $\Vec{b}$ and combine it into the saliency map. Specifically, our model learns a total of $R$ Gaussian prior maps from the data to model the center bias, whose means ($\mu_x, \mu_y$) and variances ($\sigma_x^2, \sigma_y^2$) are learned as follows:
\begin{align}
f_{\text{gau}}(x, y) = \frac{1}{2\pi\sigma_x\sigma_y} \exp(-(\frac{(x-\mu_x)^2}{2\sigma_x^2}+\frac{(y-\mu_y)^2}{2\sigma_y^2})).
\end{align}

\emph{Saliency Map Generation.} Consequently, the saliency maps are constructed by concatenating spatial feature map $\Vec{m}_e$, baseline feature map $\Vec{m}_b$ and prior maps $\Vec{b}$:
\begin{align}
    \hat{y} = f_{\text{end}}(\Vec{m}_e \mathbin\Vert \Vec{m}_b \mathbin\Vert \Vec{b}),
\end{align}
where $\mathbin\Vert$ denotes the concatenation operations and $f_{\text{end}}$ represents two convolution layers and one bilinear upsampling layer.

In the rest of this section, we describe two key components of the architecture.

\subsection{Semantic Proximity Network}

\begin{figure}[t]
\centering
\includegraphics[width=1\linewidth]{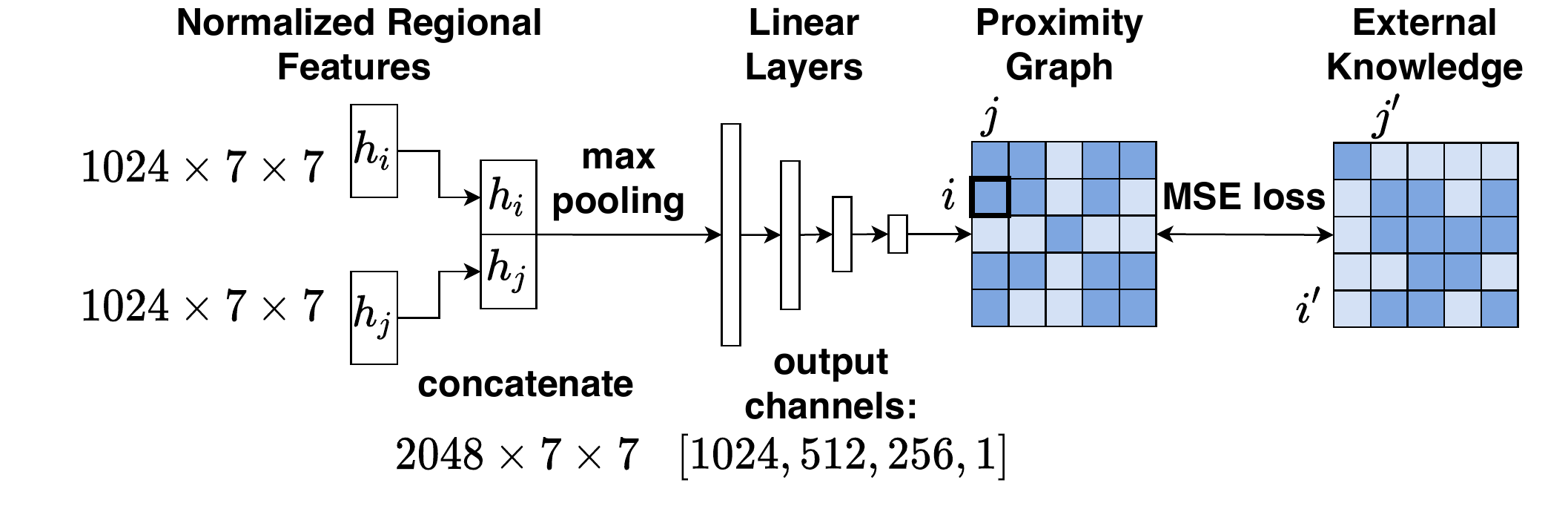}
\caption{\textit{Learning of the Semantic Proximity Network.} Region-to-region semantic proximity values are predicted by feeding concatenated regional features to a four-layer MLP. The weights of SPN are trained under the supervision of external object relationship knowledge with a MSE loss}
\label{fig:EKM1}
\vspace{-0.5em}
\end{figure}

A key component of our model is the explicit modeling of semantic proximity, with the supervision from external knowledge. As shown in \fig~\ref{fig:EKM1}, the computation of an external proximity graph consists of two steps. First, we propose a Semantic Proximity Network (SPN) that predicts the semantic proximity graph based on the input features. Each node in the semantic proximity graph represents a detected object, and the edges indicate their pairwise semantic proximity. Next, we build an external knowledge graph from semantic databases (\eg~MSCOCO captions,  WordNet), which models the semantic proximity between different object categories. While the external knowledge graph is used as the explicit supervision of the SPN, the distillation is not forced. Instead, with the distillation loss as a part of the model objective, the model can learn various semantic relationships, and how to incorporate such information is dependent on the input image. To include richer semantic proximity information, multiple knowledge graphs can be incorporated with different SPNs.

We define the $l$-th semantic proximity graph as a $p\times p$ adjacency matrix $\Mat{\hat{A}_l}$, and $\hat{e}_{ij}$ represents the learned edge connectivity between region $i$ and region $j$, where $p$ is the number of regions. The SPN aims to predict the edge connectivity between every two regions. Specifically, as shown in \fig~\ref{fig:EKM1}, the edge $\hat{e}_{ij}$ of specific graph's adjacency matrix $\Mat{\hat{A}}_l$ can be computed with a Multi-Layer Perceptron (MLP):
\begin{align}
\hat{e}_{ij} = \text{MLP}_{\Mat{\hat{A}}_l}(h_i \mathbin\Vert h_j),
\end{align}
where $\mathbin\Vert$ represents the concatenation operation. If $\hat{e}_{ij}$ is greater than a threshold $\theta_{l}$, an edge is formed between region $i$ and region $j$. %We can control the degree of proximity by selecting different $\theta_{l}$ for knowledge graph $l$.

% \subsubsection{Building External Knowledge Graphs.}
To supervise the learning of each SPN, we construct external knowledge graphs of semantic proximity. The proximity information can be obtained from multiple sources. Details about building external knowledge graphs will be discussed in Implementation Details (Section~\ref{sec:impl_details}).

\subsection{Spatial Graph Attention Network}

% To extract features from the computed semantic proximity graph for saliency prediction, we propose a Spatial Graph Attention Network (sGAT) by extending the Graph Attention Network~\cite{velivckovic2017graph}. 
We propose a Spatial Graph Attention Network (sGAT) to use the distilled external knowledge (\ie~semantic proximity graph) for saliency prediction. The sGAT is composed of multiple graph convolutional layers. The inputs to the sGAT are the regional features $\Vec{h} = \{h_1, h_2, \cdots, h_p\}$, while its output is a group of updated regional features $\Vec{h}_l' = \{h'_{l1}, h'_{l2}, \cdots, h'_{lp}\}$, $h'_{li} \in \mathbb{R}^{d_1\times d_2 \times d_3}$. The sGAT computes attention coefficients $c_{ij}$, where $i, j$ are the indices of the regions.% and $c_{ij}$ denotes the importance of region $i$'s features to region $j$'s features.}

% Since our task is to predict a saliency map, simply fattening the region features for linear transformation in Graph Attention Network~\cite{velivckovic2017graph} can destroy the hidden spatial structure. Therefore, we replace the fully-connected layers in GAT with convolutional layers to process the regional features. 

To predict saliency, it is important to preserve the spatial characteristics of the region proposals. Therefore, different from the standard GAT method, in this work $c_{ij}$ is computed as
\begin{align}
c_{ij} = f_{\text{att}}(\Mat{W}\circ h_i, \Mat{W}\circ h_j),
\end{align}
where $\Mat{W}$ are learnable weights of a spatial filter, $\circ$ denotes the convolution operation, and $f_{\text{att}}$ 
% is a shared attention mechanism: $R^{d_1\times d_2\times d_3}\times R^{d_1\times d_2\times d_3} \rightarrow R$. In this model, $f_{\text{att}}$ 
represents an attention block following GAT~\cite{velivckovic2017graph}. 
% Note that currently, attention coefficients of all possible node pairs are computed. 
The sGAT computes the coefficients of the region $i$'s immediate neighbors (include $i$) in the predicted semantic proximity graph. With a softmax normalization on $c_{ij}$, we obtain attention values $\{\alpha_{ij}\}$, indicating region $j$'s importance towards region $i$.

Finally, we obtain an updated node features $h'_{li}$ by linearly combining convoluted features from node $i$'s neighboring nodes with attention values as weights. We adopted a multi-head strategy to stabilize the learning process:
\begin{align}
h'_{li} = \mathbin\Vert_{k=1}^{K} \sigma(\sum_{j\in N_i} \alpha_{ij}\Mat{W}\circ h_j),
\label{eq:singleAttention}
\end{align}
where $\mathbin\Vert$ represents concatenation and $K=8$ is the number of attention heads.
% Similarly, to maintain the spatial structure of region features $\Vec{h}'_l$, we replace fully-connected layers with convolutional layers.

%We also inherit the multi-head strategy to stabilize the learning process. Assume $K$ independent attention mechanisms are processed, the multi-headed version of new node features in Equation~\ref{eq:singleAttention} is
% \begin{align}
% h'_i = \mathbin\Vert_{k=1}^{K} \sigma(\sum_{j\in N_i} \alpha^k_{ij}\Mat{W}^k\circ h_j).
% \end{align}
% Finally, we recoverscale all the updated region features back to its original size with biliear upsampling, and replace the original regional features in the features maps with the updated features. The updated feature maps $\Vec{m}_e$ are concatenated with the original feature maps $\Vec{m}_b$ to compute the saliency map.

\subsection{Objective}

In this work, we aim to jointly optimize the saliency prediction and the prediction of semantic proximity graph. Therefore, our model is optimized with two objective functions: the saliency prediction loss and the semantic proximity loss.

For the saliency prediction, our model is trained with a linear combination of $L_1$ loss following~\cite{yang2019dilated}) and two of the most recommended saliency evaluation metrics~\cite{bylinskii2018different} CC and NSS. They complement each other and together ensure the model's overall performance:
\begin{align}
L_{\text{sal}} = L_1(\hat{y}, y) - \beta L_{CC}(\hat{y}, y) - \gamma L_{NSS}(\hat{y}, y),
\end{align}
where $\hat{y}$ denotes the output saliency map and the ground truth is denoted as $y$.

To supervise the predicted semantic proximity graph $\hat{\Mat{A}}_l$, we need to leverage instance labels in the training phase. Assume we are going to predict the connectivity between region $i$ and region $j$, we can find their corresponding class $i'$ and $j'$ based on the positions. Next, we retrieve the corresponding ground truth edge connectivity $e_{i'j'}$ from $\Mat{A}_l$. Note that $\hat{\Mat{A}}_l$ is in $m\times m$, while $\Mat{A}_l$ is in $p\times p$, where $m$ is the number of proposed regions and $p$ is the number of classes from external knowledge. We generate multiple proximity graphs with different semantics with the mean squared error (MSE) loss:
\begin{align}
L_{\text{prox}}=\sum_{0<=i<j<p} (\hat{e}_{ij}-e_{i'j'})^2.
\end{align}
The final loss is the linear combination between saliency prediction loss and semantic proximity loss:
\begin{align}
L = L_{\text{sal}} + \lambda L_{\text{prox}}.
\end{align}

%------------------------------ Methodology Ends Here -------------------------------------------------%

\section{Experiments}
This section reports extensive comparative experiments and analyses. We first introduce the datasets, evaluation metrics, and implementation details. Next, we quantitatively compare our proposed method to the state-of-the-art saliency prediction methods. Finally, we conduct ablation studies
to examine the effect of each proposed component, and present qualitative results.

\subsection{Saliency Datasets}
We evaluate our models on four public saliency datasets: %: SALICON~\cite{jiang2015salicon}, \\ MIT1003~\cite{judd2009learning}, CAT2000~\cite{mit-saliency-benchmark} and OSIE~\cite{xu2014predicting}.
\textbf{SALICON}~\cite{jiang2015salicon} is the largest available dataset for saliency prediction. It contains 10,000 training images, 5,000 validation images and 5,000 testing images, all selected from the MSCOCO dataset~\cite{lin2014microsoft}. It provides ground-truth fixation maps by simulating eye-tracking with mouse movements. 
\textbf{MIT1003}~\cite{judd2009learning} includes 1,003 natural indoor and outdoor scenes, with 779 landscape and 228 portrait images. The eye fixations are collected from 15 observers aged between 18 to 35.
\textbf{CAT2000}~\cite{mit-saliency-benchmark} contains 4,000 images from 20 different categories, which are collected from a total of 120 observers. The dataset is divided into two sets, with 2,000 images in the training set and the rest in the test set.
\textbf{OSIE}~\cite{xu2014predicting} consists of 700 indoor and outdoor scenes from Flickr and Google, with fixations collected from 15 observers between 18 to 30. The dataset has a total of 5,551 segmented objects with fine contours.

\subsection{Evaluation Metrics}
Metrics to evaluate saliency prediction performance can be classified into two categories: distribution-based metrics and location-based metrics~\cite{bylinskii2018different,riche2013saliency}. 

We evaluate saliency models with three location-based metrics. One of the most universally accepted location-based metrics is the Area Under the ROC curve (AUC)~\cite{judd2009learning}, which treats each pixel at the saliency map as a classification task. 
% In this work, our implementation of the AUC metric follows Judd~\etal. %A variant of AUC-Judd is AUC-Bori~\cite{mit-saliency-benchmark}, which utilizes a random uniform samples of image pixels as negatives. The false positve rate (FP rate) is defined as the ratio of false positive at these sampled negatives pixels.
To take into account the center bias in eye fixations, we also use the shuffled AUC (sAUC)~\cite{zhang2008sun} that draws negative samples from fixations in other images. Another widely used metric is Normalized Scanpath Saliency (NSS)~\cite{peters2005components}, which is computed as the average normalized saliency at fixated locations. 

We also use three distribution-based metrics for model evaluation. 
% For the distribution-based metrics, the most commonly used 
One is the Linear Correlation Coefficient
(CC)~\cite{bylinskii2018different}. The CC metric is computed by dividing the covariance between predicted and ground-truth saliency map with ground truth. Besides, the similarity metric (SIM)~\cite{bylinskii2018different} is also adopted to measure the similarity between two distributions. 
% To calculate SIM, both the predicted maps and the fixation maps are normalized, followed by the summation of the min values between both maps at each pixel. 
We also computed Kullback-Leibler divergence (KL)~\cite{bylinskii2018different} that measures the difference between two distributions from the perspective of information theory. 

\setlength{\tabcolsep}{0.25em}
\begin{table}[t]
\begin{center}
\caption{\textit{Evaluation results of the the compared models.} {\color{red}RED} and {\color{blue}BLUE} indicate the best performance and the second best. The proposed model is compared with six state-of-the-art models on SALICON, MIT1003, CAT2000 and OSIE datasets under six evaluation metrics. Respectively, \model+CB and \model~denote the model with and without prior map generation as center bias}\label{table:mscoco}\label{tab:my_label}
\resizebox{1\linewidth}{!}{
\begin{tabular}{l cccccc c cccccc}
\toprule
& \multicolumn{6}{c}{SALICON} && \multicolumn{6}{c}{MIT1003} \\
\midrule
Methods & CC & AUC & NSS & sAUC & KL & SIM &$\quad$& CC & AUC & NSS & sAUC & KL & SIM \\
\midrule
% \model & \textbf{\color{red}0.867} & \textbf{\color{red}0.888} & \textbf{\color{red}3.255} & \textbf{\color{red}0.784} & \textbf{\color{red}0.598} & \textbf{\color{blue}0.801}
% &&\textbf{\color{red}0.772}&	0.909&	\textbf{\color{red}2.897}&	\textbf{\color{red}0.641}&	\textbf{\color{red}0.653}&	\textbf{\color{red}0.577} \\
\model+CB & {\color{blue}\textbf{0.866}} & \textbf{\color{red}0.892} & \textbf{\color{red}3.292} & \textbf{\color{blue}0.784} & \textbf{\color{blue}0.604} & \textbf{\color{red}0.812} &$\quad$& \textbf{\color{red}0.775} & \textbf{\color{blue} 0.910} & \textbf{\color{red}2.921} & 0.629 & \textbf{\color{red}0.574} & \textbf{\color{red}0.595} \\
\model & {\color{red}\textbf{0.867}} & \textbf{\color{blue}0.888} & \textbf{\color{blue}3.261} & \textbf{\color{red}0.786} & \textbf{\color{red}0.598} & \textbf{\color{blue}0.805} &$\quad$& \textbf{\color{blue}0.772} & 0.909 & \textbf{\color{blue}2.897} & \textbf{\color{red}0.641} & \textbf{\color{blue}0.633} & \textbf{\color{blue}0.577} \\
DINet~\cite{yang2019dilated} & 0.860 & 0.884 & 3.249 & 0.782 & 0.613 & 0.804
&& 0.764&	0.907&	2.851&	\textbf{\color{blue}0.635} &	0.690 &0.561 \\

SAM-Res~\cite{cornia2018predicting} & 0.842 & 0.883 & 3.204 & 0.779 & 0.607 &0.791 
&& 0.768&	\textbf{\color{red}0.913}&	2.893&	0.617 &0.684	&0.543 \\

SAM-VGG~\cite{cornia2018predicting} & 0.825 & 0.881 & 3.143 & 0.774 & 0.610 & 0.793 
&& 0.757&	\textbf{\color{blue}0.910}&	2.852&	0.613 &0.676	&0.568 \\

DSCLRCN~\cite{liu2018deep} & 0.831 & 0.884 & 3.157 & 0.776 & 0.637 & 0.731  
&& 0.749&	0.882&	2.817&	0.621&	0.727&	0.527 \\

SalNet~\cite{pan2016shallow} & 0.730	&0.862	&2.767	&0.731	&0.674	&0.716
&&0.727	&0.879	&2.697	&0.628	&0.763	&0.544 \\

SALICON~\cite{huang2015salicon} & 0.657 & 0.837 & 2.917 & 0.710 & 0.658 & 0.662 
&& 0.724&	0.875&	2.764&	0.613&	0.818&	0.534 \\
\toprule
& \multicolumn{6}{c}{CAT2000} && \multicolumn{6}{c}{OSIE} \\
\midrule
% \model & \textbf{\color{blue}0.891}&	\textbf{\color{red}0.884}&	\textbf{\color{red}2.387}&	\textbf{\color{red}0.617}&	0.564 &\textbf{\color{blue}0.770 }
% &&\textbf{\color{red}0.849}	&\textbf{\color{red}0.911}	&\textbf{\color{red}3.494}	&\textbf{\color{red}0.875}	&\textbf{\color{red}0.731}	&\textbf{\color{red}0.716} \\
\model+CB & \textbf{\color{red}0.897} & \textbf{\color{red}0.889} & \textbf{\color{red}2.481} & 0.610 & \textbf{\color{red}0.529} & \textbf{\color{red}0.785} &$\quad$& \textbf{\color{red}0.853} & \textbf{\color{red}0.911} & \textbf{\color{red}3.324} & 0.859 & \textbf{\color{blue}0.711} & \textbf{\color{red}0.725} \\
\model & \textbf{\color{blue}0.894} & \textbf{\color{blue}0.886} & \textbf{\color{blue}2.413} & \textbf{\color{red}0.617} & 0.567 & \textbf{\color{blue}0.779} &$\quad$& \textbf{\color{blue}0.847} & \textbf{\color{blue}0.906} & \textbf{\color{blue}3.317} & \textbf{\color{red}0.864} & 0.729 & \textbf{\color{blue}0.724} \\

DINet~\cite{yang2019dilated} &0.874 &0.877 &2.379	&\textbf{\color{blue}0.612}	&0.598	&0.765 && 0.842 & 0.903 & 3.264 & 0.860 &0.751 & 0.718 \\

SAM-Res~\cite{cornia2018predicting} & 0.892	&0.883 &2.386	&0.585	&0.563	&0.778 
&& 0.843 & 0.901 & 3.237 & \textbf{\color{blue}0.862} & \textbf{\color{red}0.704} & 0.723 \\

SAM-VGG~\cite{cornia2018predicting} &0.891	&0.882	&2.387	&0.581	&\textbf{\color{blue}0.547}	&0.762 
&& 0.832 & 0.893 & 3.196 & 0.858 & 0.727 & 0.690 \\

DSCLRCN~\cite{liu2018deep} &0.834	&0.861	&2.357	&0.541	&0.851	&0.684 
&& 0.667 & 0.882 & 2.621 & 0.831 & 0.878 & 0.499 \\

SalNet~\cite{pan2016shallow} &0.817	&0.864	&2.361	&0.563	&0.674	&0.663
&& 0.805 & 0.887 & 2.897 & 0.837 & 0.764 & 0.624 \\

SALICON~\cite{huang2015salicon} & 0.801	& 0.862	& 2.343	& 0.524	& 0.867	& 0.652 
&& 0.686 & 0.890 & 2.849 & 0.842 & 0.725 & 0.566 \\
\bottomrule
\end{tabular}
}
\end{center}
\vspace{-0.5em}
\end{table}
% ----------------------------------------------------------------------

\subsection{Implementation Details}
\label{sec:impl_details}
Our CNN backbone follows the design of DINet~\cite{yang2019dilated}, which is a dilated ResNet-50 network with the convolution layers in the last two blocks replaced with dilated convolution. The parameters of the backbone are initialized using a ResNet-50 network pre-trained on ImageNet~\cite{krizhevsky2012imagenet}.

The Faster~R-CNN object detector is trained on the MSCOCO dataset with default hyper-parameters~\cite{ren2015faster}. Instead of using the original anchor size, we adopt a small anchor size $\{64, 128, 256\}$, which shows a better performance in detecting small objects in the scene. 
% Besides, as removing SALICON images from MSCOCO has only minor impacts over the model performance, we doesn't take this into consideration.
%The extracted regional features are fed into a four-layer MLP to predict two semantic proximity graphs under the supervision of external class-to-class proximity information from MSCOCO captions and WordNet. 

We consider two external knowledge sources for building the ground-truth semantic proximity graph: MSCOCO image captioning and WordNet, learned with two SPNs. For the MSCOCO image captioning data, if different semantics (\ie~MSCOCO object categories, for simplicity) are frequently mentioned in the captions of different images, we consider them to be close to each other in the semantic space. Specifically, we use the number of occurrence between two semantics divided by the max occurrence as the value of the entry. For the WordNet data, we can retrieve the wup\_similarity value~\cite{wu1994verbs} between every pair of object categories from the WordNet to produce a ground-truth semantic proximity graph. The thresholds to identify a predicted edge from the SPNs are $0.3$ for MSCOCO image captioning and $0.5$ for WordNet.

%Here we define the number of images $\theta_{cap}$ as the threshold value to separate correlated and uncorrelated pairs. 
% assume the adjacency matrix for MSCOCO captions is $\Mat{A}_{\text{cap}}$, 

% With these measures, we can build two different external knowledge graphs about semantic proximity, which will be used to supervise the Semantic Proximity Network (SPN).

%Next, the normalized regional features are convoluted across these proximity graphs with sGAT, producing multiple updated regional proximity features. These features are recovered into their original size and replaced the old ones, producing proximity feature map with 256 channels for each knowledge source. Besides, the backbone features are also fed into baseline saliency network, whose output are concatenate with proximity feature map to predict saliency map.

In our experiments, we train and evaluate our model with and without modeling the center bias. We set the number of prior maps $R=16$. For the SALICON dataset, we train the model on its training set and evaluate it on the validation set. The size of mini-batch is 10 and the optimizer is Adam optimizer~\cite{kingma2014adam}. The initial learning rate is $10^{-3}$ and the learning decay rate is $10^{-4}$. For the other datasets, we fine-tuned the model trained on SALICON with the corresponding eye-tracking data. We randomly select 80\% of the samples for training and use the rest 20\% for validation. During the fine-tuning, the size of mini-batch is 10, and the optimizer is Adam optimizer. The initial learning rate is decreased to $10^{-4}$ and the learning decay rate is $10^{-4}$. To ensure a fair comparison, we replicated the compared models using the same training and validation sets as ours.

% details about how to build external knowledge graph

% -----------------------------------------------------------------------------------------

% \subsection{Dilated Residual Module}
% One of the commonly pre-trained backbone networks in saliency prediction is ResNet-50 with shortcut connections~\cite{he2015convolutional}. However, with the increased model depth, more down-sampling operations are performed, contributing to a decreased output resolution. To handle the issue of limited output spatial resolution, we use dilated ResNet-50 from SAM-Res~\cite{cornia2018predicting}.

% In particular, the dilated version of ResNet-50 network consists of five convolutional blocks and a fully connected layer. The first block is composed of one convolutional layer, followed by a maxpooling layer with a stride of 2. Conv2 and Conv3 are left intact. The standard convolution operations in the remaining two blocks are replaced by dilated convolution, with dilation rates of 2 and 4, respectively. Besides, in Conv4 and Conv5, downsample operations are removed. The last fully connected layer is also omitted to extract spatial image features for the rest modules.

\begin{figure}[t]
\centering
\includegraphics[width=\linewidth]{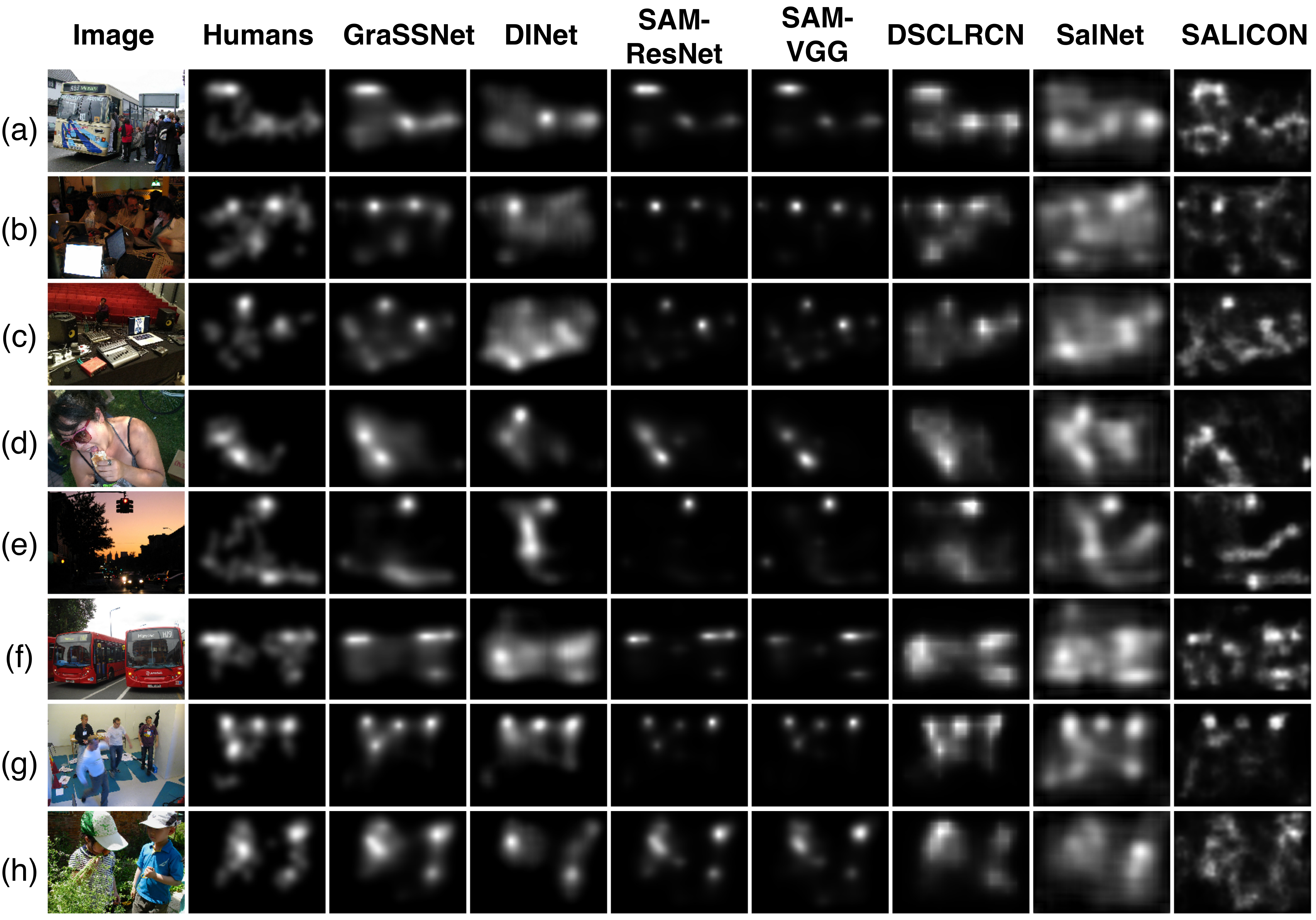}
\caption{\textit{Qualitative comparison between our model and state-of-the-art saliency prediction methods.} In each row, we list the image, ground truth, saliency maps without prior maps of our method and six state-of-the-arts models. Examples (a)-(e) demonstrate images with different object categories and examples (f)-(h) demonstrate images with the same object categories}
\label{fig:qualitive}
\vspace{-0.5em}
\end{figure}

\subsection{Quantitative Analysis}
% We quantitatively compare our method with the state of the
% art on SALICON~\cite{}, OSIE~\cite{}, MIT1003~\cite{} and CAT2000~\cite{} datasets.
As shown in \tab~\ref{tab:my_label}, our method achieves state-of-the-art performances on all the datasets. It consistently outperforms other methods in all the metrics on SALICON.
The promising results suggest that modeling semantic proximity is effective for improving the overall performance of saliency prediction. On the other datasets, \model~also achieves better performances than the DINet that shares the same backbone as ours. Also, due to the differences in image characteristics among these datasets, the promising results demonstrate that the knowledge learned from the SALICON data can be successfully transferred to the other saliency datasets.  It is noteworthy that \model~includes CC and NSS as part of the objective, which gives advantages on CAT2000, MIT1003 and OSIE under both metrics. They complement each other and together ensure the model’s overall performance. The SIM scores of our model are also significantly better than the others even though SIM is not used as a training loss. Similarly, the KL scores are the best on SALICON, MIT1003, CAT2000 and the second-best on OSIE. Besides, our model also maintains improved sAUC values over DINet on SALICON when center bias is explicitly modeled.

\begin{figure}[t]
    \centering
    \includegraphics[width=\linewidth]{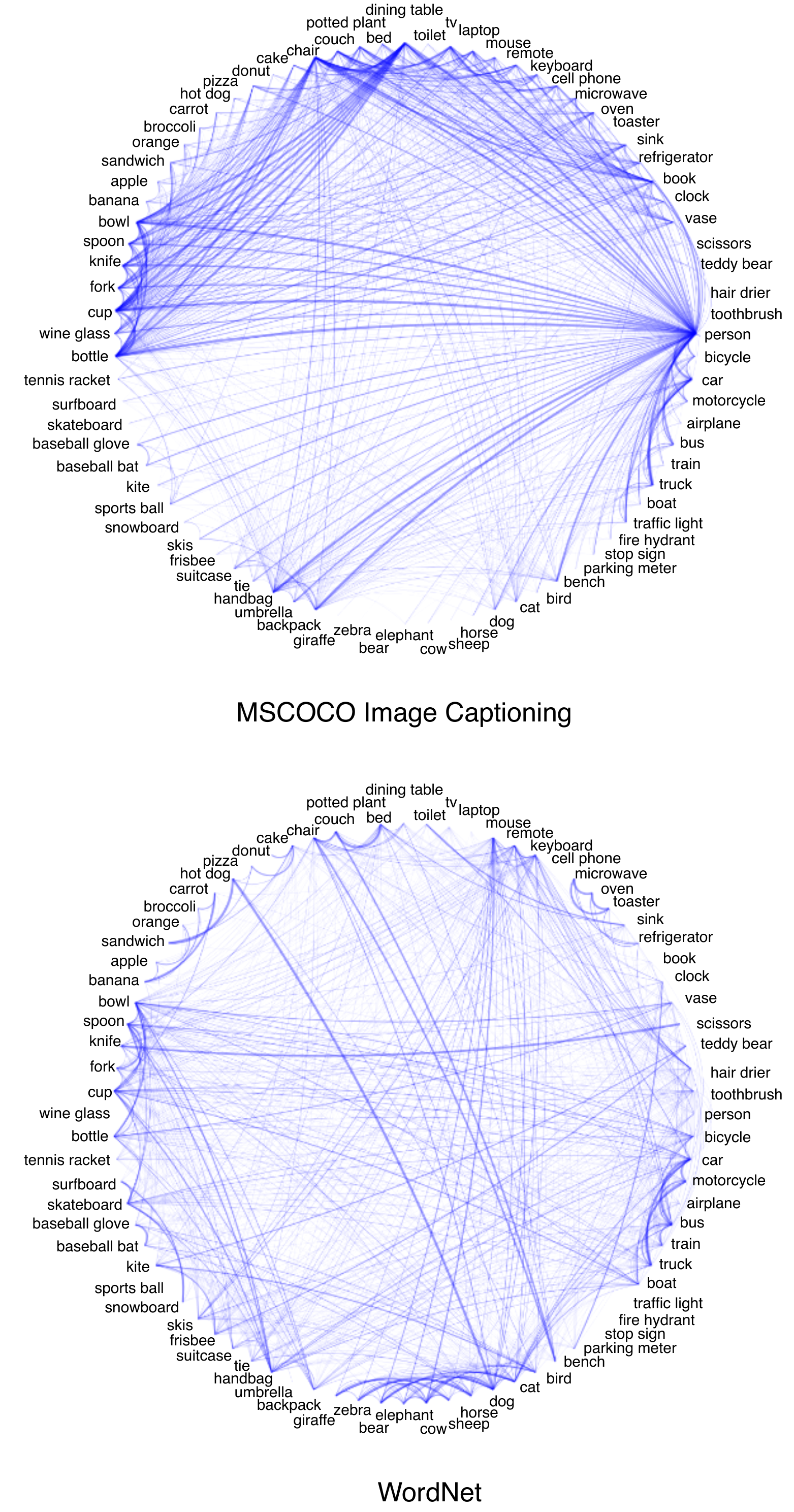}
    \caption{\textit{Visualization of external semantic proximity graphs.} We visualize knowledge graph from MSCOCO image captioning (left) and WordNet (right) with classes as nodes in cycle layout. The width of the link indicates the strength of semantic proximity between the connected classes}
    \label{fig:knowledge_graph}
\vspace{-0.5em}
\end{figure}

\subsection{Qualitative Analysis}
We report the qualitative results of our model, in comparison with the state-of-the-art approaches. These qualitative examples are selected from the SALICON validation set, which demonstrates complex scenes in which semantic proximity can effectively improve saliency prediction. As shown in \fig~\ref{fig:qualitive}, our \model~method performs the best for complex scenes with many objects in the foreground and background. In particular, for salient objects with strong semantic relationships, including both different objects (\eg~people and computers in \fig~\ref{fig:qualitive}c) and similar objects (\eg~multiple people in \fig~\ref{fig:qualitive}g), our method successfully predicts the relative saliency among these salient regions.  To be more specific, taking \fig~\ref{fig:qualitive}a-d for example, our method captures the close relationship between the person and bus/computer/food, and hence highlights both. Similarly, both the traffic light and the car (\fig~\ref{fig:qualitive}e) get highlighted due to their strong semantic relationship. Besides, as can be seen from (\fig~\ref{fig:qualitive}f-h), which consists of multiple buses/people, the features are interchanged among them to intensify the saliency. A more detailed analysis of how regions are connected to produce proximity graph and how the distilled information benefits saliency prediction are discussed in Section~\ref{sec:visualization}.

%Moreover, as we predict region-to-region connectivity, these objects can either be in different types or in the same types. Take the images of human and bread for example, region of human type can propagate its salient features to bread, increasing the saliency values in the bread region. Also, example of two giraffes demonstrates how the object of the same type can share saliency features with each other.

% To sum up, the proposed model exhibits capability in utilizing the external semantic proximity information to produce better performance.

\subsection{Ablation Studies}
To demonstrate the effectiveness of the various components and hyper-parameters used in the model, we conduct ablation studies on the SALICON dataset.
% The ablation analysis includes the evaluation of model components, the influences of choosing different bounding box width and backbone network. 

\begin{figure*}[t]
    \centering
    \includegraphics[width=\linewidth]{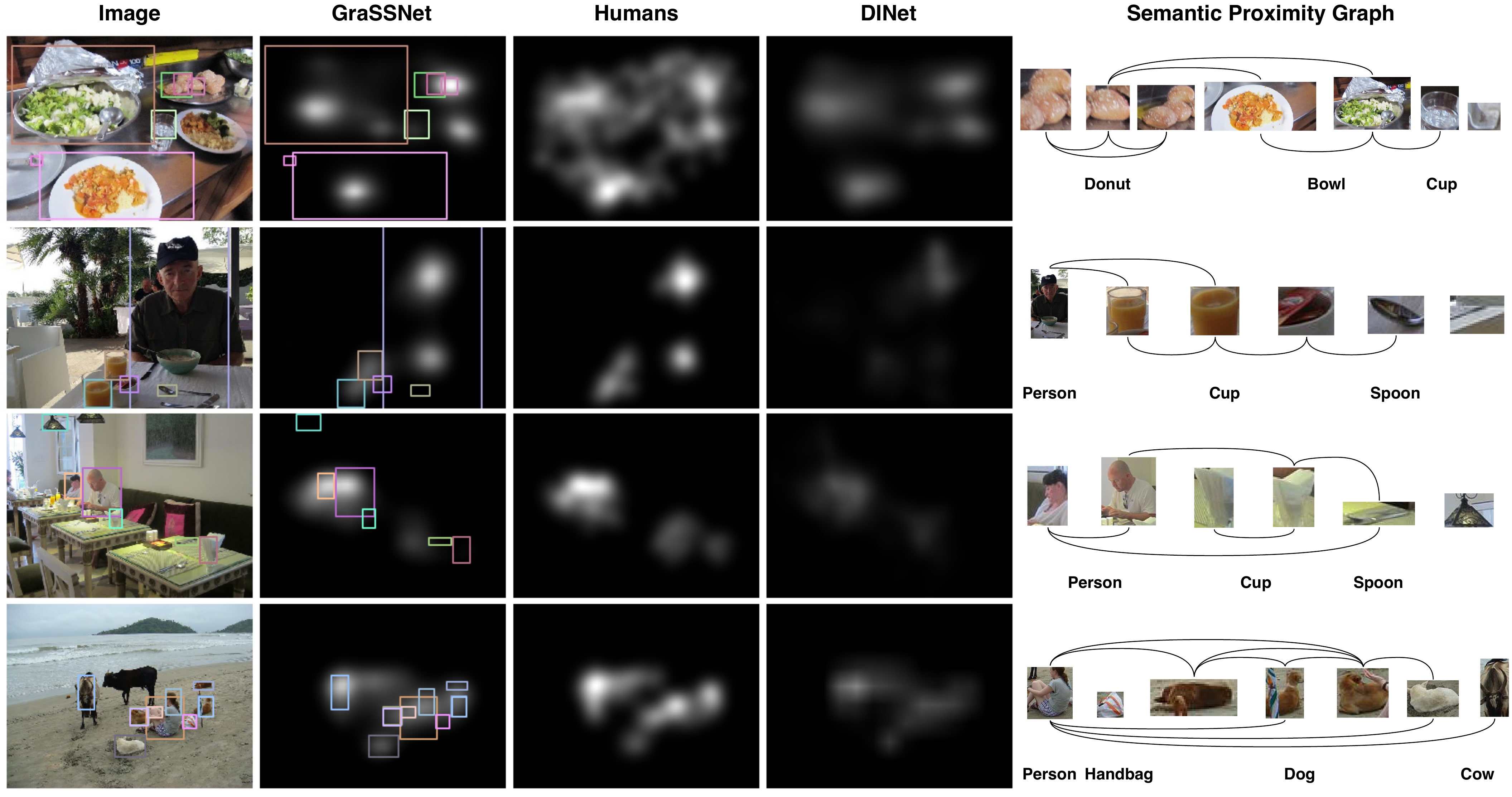}
    \caption{\textit{Visualization of region proposals and semantic proximity graphs.} We show the proposed bounding boxes (the first column), saliency prediction results of our method without prior maps (the second column) and connected the regions with predicted edges from SPN (the last column). Besides, we list the ground truth fixation maps (the third column) and the saliency map of DINet (the fourth column)}
    \label{fig:ablation_vis}
\vspace{-0.5em}
\end{figure*}

\setlength{\tabcolsep}{0.5em}
\begin{table}[t]
\begin{center}
\caption{\textit{Ablation study of the external knowledge on the SALICON dataset.} We test the performance of models without prior maps trained with different combinations of external knowledge for three different backbones (ResNet-50, ResNet-101 and VGG-19)}
\label{table:ablation_structure}
\resizebox{1\linewidth}{!}{
\begin{tabular}{cccccc}
\toprule
% \multicolumn{8}{c}{Model Structures} \\
% \midrule
Backbone & External Knowledge &CC &AUC	&NSS &sAUC \\
\midrule

ResNet-50&	-	&	                0.851&	0.874&	3.239&	0.762 \\

ResNet-50&	MSCOCO&		            0.863&	0.886&	3.251&	0.784 \\

ResNet-50& WordNet&               0.861&	0.884&	3.247&	0.778 \\

ResNet-50&	MSCOCO + WordNet&		\textbf{0.867}&	\textbf{0.888}&	\textbf{3.261}& 	\textbf{0.786}\\
\midrule

ResNet-101&		-&         0.841&	0.864&	3.084&	0.757\\      

ResNet-101&	MSCOCO&	   0.861&	0.887&	3.252&	0.777\\

ResNet-101& WordNet&       0.859&	0.879&	3.234&	0.761 \\
ResNet-101&	MSCOCO + WordNet& \textbf{0.864}&	\textbf{0.888}&	\textbf{3.254}&	\textbf{0.778}\\
\midrule
VGG-19    &		-&         0.834&	0.854&	2.915&	0.749\\

VGG-19    &	MSCOCO&	0.855&	0.882&	3.246&	0.775\\

VGG-19    & WordNet&     0.853&  0.877&  3.243& 0.772 \\

VGG-19    &	MSCOCO + WordNet& \textbf{0.859}&	\textbf{0.886}&	\textbf{3.249}&	\textbf{0.776}\\

\bottomrule
\end{tabular}
}
\end{center}
\vspace{-0.5em}
\end{table}

\subsubsection{Effects of External Knowledge}
We first examine how external knowledge benefits the saliency prediction. \tab~\ref{table:ablation_structure} reports the ablation study on the incorporation of external knowledge. On three different backbone networks, the comparison between models with and without external knowledge supervision shows that inclusion of external knowledge from MSCOCO image captioning and WordNet both improve the model performance. The results suggest that external knowledge about semantic proximity from both data sources can provide essential information for saliency prediction. \fig~\ref{fig:knowledge_graph} visualizes the semantic proximity graphs built from MSCOCO image captioning and WordNet, where nodes represent the 80 MSCOCO categories and the width of edges represents proximity. The figure illustrates that knowledge in MSCOCO is human-centric, providing a list of classes that commonly occur at the presence of the person type (\eg~handbag, spoon, cup, bowl, \etc). Besides, the knowledge graph from WordNet is relevant to the taxonomy of object types. It is quite effective in saliency prediction because objects of the same class are likely to appear together (\eg knife, fork and spoon often present together as dinnerware). Taking multiple knowledge bases into account is helpful for the model's generalizability to a broader domain.

%For example, the performance of the complete structure (DINet + sGAT + RPN + External Knowledge) is better than that without external semantic proximity knowledge (DINet) in terms of CC, AUC, NSS and sAUC. It need to be mentioned that we do not test RPN, sGAT and external knowledge individually because they function as a united structure. It also needs to be noted that including decent backbone saliency network can benefit performance, as the backbone provides essential multi-scale internal features for saliency prediction. 

%  ------------------------------------------------------------------------------------
% differences in the backbone network ((ResNet-50 \vs ResNet-101)) does not affect the performance much, which suggests that ResNet-50 model has encoded sufficient semantic features for saliency prediction.
% -------------------------------------------------------------------------------------------

% moved to supp

% ---------------------------------------------------------------------------------
\subsubsection{Effects of Hyper-Parameters}
Here we examine the choice of hyper-parameters. Firstly, since our loss function is a linear combination of $L_1$ distance, CC and NSS, as well as
the MSE loss of edge predictions, we explore how different combinations of the parameters $\{\beta, \gamma,\lambda\}$ influence model performance. Results from \tab~\ref{table:ablation_param} indicate that setting $\beta=0.3$, $\gamma=0.15$ and $\lambda=0.8$ can optimally balance the scores of different evaluation metrics and achieve the overall best performance.
% -------------------------------------------------------------------------------------------
\begin{table}[t]
\begin{center}
\caption{\textit{Ablation study of the hyper-parameters on the SALICON dataset.} We report model performances without prior maps in four metrics (CC, AUC, NSS and sAUC) under different combinations of $\beta$, $\gamma$, $\lambda$}
\label{table:ablation_param}
% \resizebox{0.6\textwidth}{!}{
\begin{tabular}{ccccccc}
\toprule
% \multicolumn{6}{c}{Other Parameters} \\
% \midrule
$\beta$ &$\gamma$ &$\lambda$ &CC &AUC	&NSS &sAUC \\
\midrule
% the optimal values for parameter
0.3 & 0.15 & 0.8 & \textbf{0.867} & \textbf{0.888} & \textbf{3.261} & \textbf{0.786} \\
\midrule
0.01 & 0.15 & 0.8 & 0.848	&0.880	&3.248	&0.782 \\
0.1 & 0.15 & 0.8 & 0.864	&0.887	&3.254	&0.785 \\
1 & 0.15 & 0.8 & 0.857	& 0.882	& 3.241	& 0.774 \\
10 & 0.15 & 0.8 & 0.85	& 0.876	& 3.227	& 0.769 \\
\midrule
0.3 & 0.01 & 0.8 & 0.858 &0.873	& 3.227	& 0.776 \\
0.3 & 0.1 & 0.8 & 0.862	&0.881	&3.242	&0.781 \\
0.3 & 1 & 0.8 & 0.843 & 0.879	&3.255	&0.773 \\
0.3 & 10 & 0.8 & 0.837&	0.868	&3.240	&0.756 \\
\midrule
0.3 & 0.15 & 0.1 & 0.841 & 0.870 & 3.236 &0.770 \\
0.3 & 0.15 & 1 & 0.861 & 0.884 & 3.250 & 0.778 \\
0.3 & 0.15 & 10 & 0.839	& 0.868	& 3.212	& 0.772 \\
% 6&        	    0.1&		0.1&    0.851&	0.875&	3.231&	0.775 \\
% 8&        	    0.1&	    0.1&	0.867&  0.888&  3.255&  0.784 \\

% 10&       	    0.1&		0.1&    0.864&	0.881&	3.244&	0.781 \\

% 12&       	    0.1&		0.1&    0.859&	0.874&	3.241&	0.775 \\

% 8&         0.001&   0.1  &	     &       &       & \\
% 8&         0.01&   0.1   &	0.861&	0.878&	3.250&	0.775 \\

% 8&             1&	0.1  &	0.865&	0.876&	3.245&	0.772 \\
% 8&             10&	0.1  &       &       &       &    \\
% \midrule
% 8&        	    0.1&		1 &	0.854&	0.872&	3.249&	0.772 \\

% 8&        	    0.1&		0.01&	0.859&	0.867&	3.237&	0.771 \\
\bottomrule
\end{tabular}
% }
\end{center}
\vspace{-0.5em}
\end{table}
% ---------------------------------------------------------------------------------
%One explanation is that the external image features help to capture objected-based features from input images and also serves as a supplement of object-based features. 
\subsubsection{Visualizations}
\label{sec:visualization}
We visualize detected object regions and predicted semantic proximity graphs in \fig~\ref{fig:ablation_vis}, to illustrate the effects of semantic proximity information on saliency prediction. As can be seen, regions of the same category or related categories are interconnected with edges. {Generally, edges are formed among all the donuts in \fig~\ref{fig:ablation_vis}a, most dogs in \fig~\ref{fig:ablation_vis}d,  cups, spoons, and bowls in \fig~\ref{fig:ablation_vis}b.} Such semantic proximity reflects the taxonomy of these words from the WordNet. Besides, some categories of objects are more related to people (\eg~handbag, cup, spoon, dog, \etc). This kind of human-centric semantic proximity is mostly derived from the MSCOCO image captioning. By taking into account the semantic proximity graphs, our model can better predict the saliency of semantically related regions.

\section{Conclusion}
In this paper, we present a novel saliency prediction network that explicitly models the semantic proximity as a graph, based on detected objects from the input. One of our key technical contributions is the novel SPN supervised by external knowledge. Beyond that, we proposed the sGAT to propagate the semantic information across the graph nodes, while preserving spatial features in node attributes.  
The modeling of semantic proximity allows our model to take the semantic relationships among multiple objects into account, and to better predict their relative saliency. The proposed method achieves promising performances on multiple saliency datasets. In future studies, we aim to extend this work by considering specific relationship modeling with the scene graph. We will also extend this work to video saliency and top-down saliency prediction.
% Besides, as the proximity graph are predicted in region-to-region manner, our model covers objects with the same and distinct classes. Apart from that, we modify the graph attention network to enable the graph convolution operation with. Finally, the effectiveness of injecting concrete external knowledge is illustrated by both quantitative and qualitative results from extensive experiments.
\newpage

% anonymous address:

{\small
\bibliographystyle{ieee_fullname}
\bibliography{refs}
}

\end{document}